\documentclass[10pt,letterpaper]{article}
\usepackage{cogsci}
\cogscifinalcopy 

\usepackage{pslatex}
\usepackage{apacite}
\usepackage{float} 

\usepackage{algorithm}
\usepackage[noend]{algorithmic}
\usepackage{graphicx}
\usepackage{multirow}

\newcommand{\code}[1]{\texttt{\small#1}}

\title{Decomposed Inductive Procedure Learning:\\Learning Academic Tasks with Human-Like Data Efficiency}

\author{
    Daniel Weitekamp,\textsuperscript{\rm 1}
    Christopher MacLellan, \textsuperscript{\rm 1}
    Erik Harpstead, \textsuperscript{\rm 2}
    Kenneth Koedinger\textsuperscript{\rm 2} \\
    \textsuperscript{\rm 1} Georgia Institute of Technology,
    Atlanta, GA 30332 \\
    \textsuperscript{\rm 2} Carnegie Mellon University,
    Pittsburgh, PA 15213
}

\begin{document}

\maketitle

\begin{abstract}
Human learning relies on specialization---distinct cognitive mechanisms working together to enable rapid learning. In contrast, most modern neural networks rely on a single mechanism: gradient descent over an objective function. This raises the question: might human learners’ relatively rapid learning from just tens of examples instead of tens of thousands in data-driven deep learning arise from our ability to use multiple specialized mechanisms of learning in combination?
We investigate this question through an ablation analysis of inductive human learning simulations in online tutoring environments. Comparing reinforcement learning to a more data-efficient 3-mechanism symbolic rule induction approach, we find that decomposing learning into multiple distinct mechanisms significantly improves data efficiency, bringing it in line with human learning. Furthermore, we show that this decomposition has a greater impact on efficiency than the distinction between symbolic and subsymbolic learning alone. Efforts to align data-driven machine learning with human learning often overlook the stark difference in learning efficiency. Our findings suggest that integrating multiple specialized learning mechanisms may be key to bridging this gap.
\end{abstract}


A key idea within the learning sciences, popularized by Anderson's ACT-R theory \citeyear{anderson2013implications} and expanded upon by others \cite{koedinger2012knowledge}, is that human performance is enabled by independent knowledge components---individual facts, skills, or principles---that must be understood and retained to exhibit mastery of higher-level capabilities. For instance, addition tables from 1 to 10 comprise $\frac{10*(10+1)}{2}=55$ facts. More complex procedures, such as adding two large numbers, may require several additional skills like aligning numbers, adding over columns, and carrying the tens-digits of partial sums. More advanced capabilities require mastery of even more interdependent knowledge components which may build upon these.

Intelligent tutoring systems (ITS) are educational technologies that mimic one-on-one tutoring interactions by providing highly adaptive step-by-step instructional support designed to aid the acquisition of unmastered knowledge components.
When students practice skills in these controlled learning environments, their rate of learning proves to be remarkably quick and astonishingly consistent between individuals \cite{koedinger2023astonishing}. Data from ITSs \cite{koedinger2010data} illustrate that students typically master skills in about a dozen practice opportunities or fewer---orders of magnitude faster than modern data-driven machine learning (ML) approaches, such as reinforcement learning, which relies on gradient-descent and often require tens of thousands to millions of examples. 

This work investigates three learning mechanisms that have emerged from efforts to simulate humans' inductive learning of academic tasks. Simulated learner systems like Sierra \cite{vanlehn1990mind}, SimStudent \cite{matsuda2015teaching}, the Apprentice Learner (AL) architecture \cite{maclellan2016apprentice}, and AI2T \cite{weitekamp2024ai2t} have successfully learned dozens of domains, acquiring skills directly from ITSs and similar environments, and in some cases, directly from human instruction. These systems have not only replicated the remarkable rate of learning observed in humans \cite{maclellan2016apprentice, weitekamp2019toward}, but also produce patterns of error that mirror those found in student data \cite{vanlehn1990mind,weitekamp2020investigating}.  

The central question of this work is: why have cognitive systems succeeded at replicating rapid human learning, while neural network approaches lag behind by several orders of magnitude? While symbolic learning mechanisms in simulated learners certainly play a role,  we show that the key factor to their efficiency is the integration of multiple functionally distinct learning systems. Through an ablation analysis, we compare a single learning mechanism (either neural reinforcement learning or a symbolic learning approach) with progressively more decomposed learning that uses two, and finally, three mechanisms, as used in prior simulated learners. We coin the name {\it Decomposed Inductive Procedure Learning (DIPL)} to refer to this common multi-mechanism approach. 

Across two ITS tasks, we show that each stage of ablation---from a single mechanism learning to DIPL's  3-mechanism learning---yields several orders of magnitude of learning efficiency improvement. We hypothesize that these gains arise from how different mechanisms, each with a well-defined role, simplify error attribution and reduce the total computational complexity of knowledge induction. This work highlights the vast benefits that theory-driven cognitive modeling can offer to the fields of cognitive science and machine learning. In an odd divergence from historical definitions \cite{vanlehn1994applications}, large language models (LLMs) have inspired a wave of so-called ``simulated learners'' that generate responses but do not actually learn \cite{kaser2024simulated}. In response, we highlight the fundamental advantage of a cognitive theory-based approach over simple gradient descent based learning in artificial neural networks. Our direct comparison between reinforcement learning and simulated learners suggests that an essential component of this advantage lies in how several specialized mechanisms of learning cooperate to achieve human-like learning efficiency.

\begin{figure}[t]
\centering
\includegraphics[width=0.99\columnwidth]{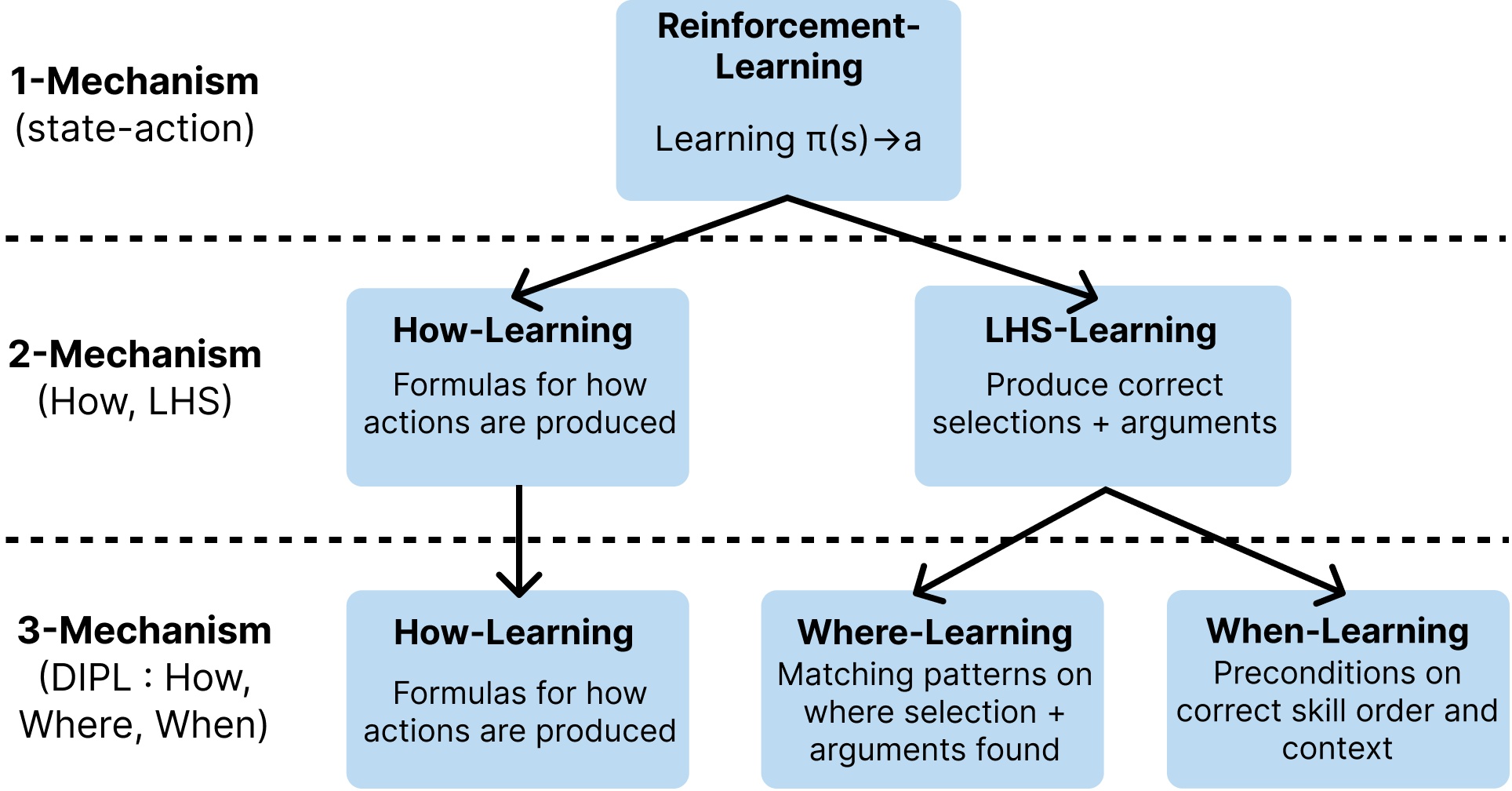} 
\caption{Decomposition from 1-mechanism learning, like RL, that maps states to actions to DIPL’s 3-mechanism learning. A 2-mechanism system bridges the difference and uses \textit{how-learning} but combines \textit{where-} and \textit{when-learning}.}
\label{decomp}
\end{figure}


\section{Related Work}

\subsection{Large Language Models}
Artificial neural systems have come a long way in performing academic tasks such as mathematics. For instance, LLMs have succeeded on challenging word problems \cite{bubeck2023sparks} from datasets like MATH  \cite{hendrycks2021measuring} and GSM8K \cite{cobbe2021training}. In 2023, GPT models trained step-by-step have demonstrated 78\% accuracy on the MATH dataset \cite{lightman2023let}, and yet higher accuracies have been achieved with several recent models on several benchmarks \cite{team2024gemini}. However, since these models were trained on a quantity of learning experiences many orders of magnitude greater than a human would experience in a lifetime, their decidedly data-driven developmental process offers little insight into human learning. Other misalignments of deep learning with human learning include the phenomena of catastrophic forgetting \cite{mccloskey1989catastrophic}, where new training causes models to forget capabilities learned from previous training instances. Additionally, strong evidence has emerged showing that LLM's problem-solving capabilities are largely memorized and not in fact a result of effective generalization from data. For instance, Mirzadeh et. al \citeyear{mirzadeh2024gsm} show that LLMs' performance drastically declines when questions from the GSM8K dataset have their numerical values replaced with different values or if additional distractor clauses are added.

\subsection{Speed-up Learning in Cognitive Systems}

A shortcoming of many attempts to model learning in symbolic cognitive systems has been the over-reliance on knowledge engineering and ``learning'' by planning with hard-coded, and often domain specifc, prior knowledge representations. Much work in logic programming \cite{manhaeve2018deepproblog} and a great deal of early work in cognitive architectures, like SOAR \cite{laird2019soar} and ACT-R \cite{ritter2019act}, can be held to this criticism. No doubt, models of learning ought to represent and build upon prior knowledge. Yet, many cognitive systems neglect important aspects of learning by starting with knowledge representations that are hard-coded to plan toward target tasks. \citeA{laird2017standard} have presented the view that learning is only ``a side effect of performance.'' This claim evokes the principle of learning-by-doing, yet is greatly misaligned with the realities of learning in an academic setting. ``Learning'' for these systems is typically characterized as a cognitive speed-up achieved by repurposing prior knowledge to shortcut later computational effort. In an academic setting, one would not consider a human student to have learned if they could ``perform'' without errors in advance of instruction. The limited focus of many cognitive systems on error-free speed-up learning \cite{neves1985learning} overlooks this important inductive component of human knowledge formation that enables the rapid, yet initially error-prone acquisition of entirely new capabilties through instruction and practice.

\section{Inductive Simulated Learners}

VanLehn’s \citeyear{vanlehn1990mind} Sierra is an early  inductive simulated learner (SL). Experiments with Sierra demonstrated that inductive learning underlies the acquisition of early mathematical skills,  replicating and explaining more mistakes in subtraction problem-solving datasets than prior case-by-case analyses. Later efforts with SimStudent and the Apprentice Learner (AL) architecture have empirically reproduced student learning curves across dozens of ITS domains \cite{maclellan2020domain}. Instead of fitting to student data, these systems learn to solve problems from the same ITS interactions human students experience, generating step-by-step solutions that improve with practice. \cite{maclellan2016apprentice,weitekamp2020investigating}. These simulated learners induce production rules from ITS examples and correctness feedback---initially producing errors, but are rapidly restructured towards mastery through supervised practice. AI2T \cite{weitekamp2024ai2t} extends this principle, letting untrained users teach it interactively. Notably, half of AI2T users successfully trained it to exhibit 100\% correct and complete behavior on mathematical tasks with small user-selected training sequences of just 14-21 problems.

\section{Decomposed Inductive Procedure Learning}

SimStudent and the Apprentice Learner induce production rules using a combination of three learning mechanisms that independently determine \textit{how} actions are taken, \textit{where} it is possible to take actions, and \textit{when} (i.e. under what circumstances and in what order) those actions should be applied to execute a target behavior. Sierra and AI2T include a fourth mechanism for inducing the hierarchical \textit{process} by which higher-level tasks are divided into subtasks. This \textit{process-learning} mechanism arranges production rules into hierarchical task networks \cite{erol1994semantics} that control how high-level tasks are broken down into partially ordered subtasks. Our notion of Decomposed Inductive Procedure Learning (DIPL) encompasses both these 3- and 4-mechanism approaches. However, for our evaluations, we focus on 3-mechanism DIPL, which includes only \textit{how-learning}, \textit{where-learning}, and \textit{when-learning}.


While multi-mechanism learning approaches, such as the actor-critic paradigm \cite{konda1999actor}, are commonplace in AI, DIPL's learning mechanisms cooperate in a uniquely modular, localized fashion. Unlike most actor-critic methods, DIPL's learning mechanisms do not apply globally; instead, each is instantiated separately for every learned skill (i.e., production rule).
Within the induction of a single skill, these mechanisms cooperate in such a way that each mechanism simplifies learning for the others. 
Rather than relying on a single global mechanism like gradient descent, each instance of each individual learning mechanism serves a distinct, well-defined role---acquiring specific generalizations for individual skills. Each skill, in turn, is built up from these induced pieces and is responsible for performing particular kinds of actions. As a model of learning, DIPL's division of capabilities into individual skills aligns with the notion of knowledge components, expressed as production rules that are refined over time within an evolving expert system. 





\subsection{How-Learning}

\begin{figure}[t]
\centering
\includegraphics[width=0.95\columnwidth]{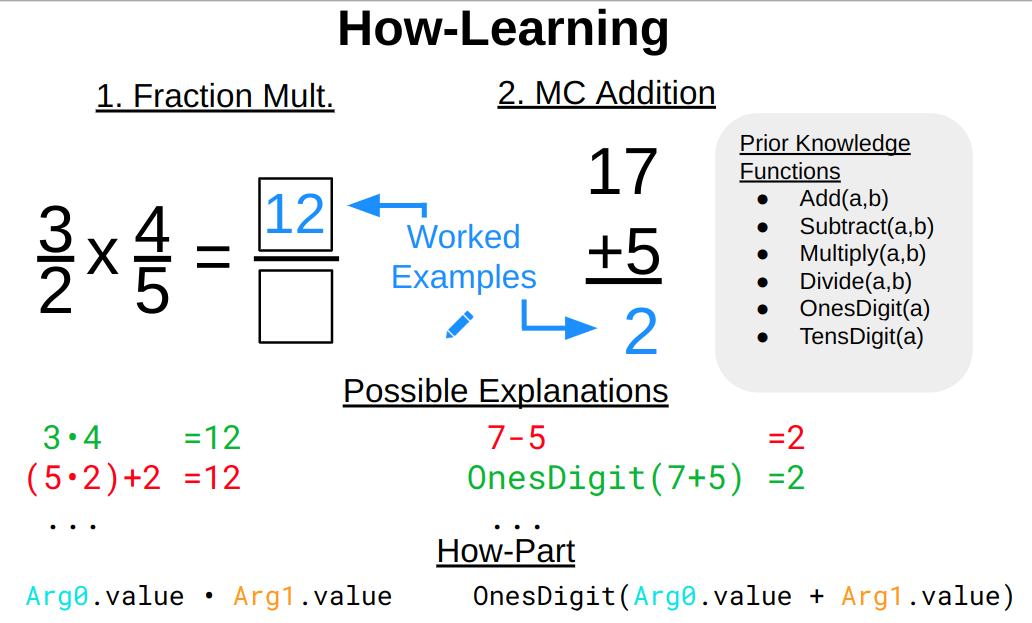} 
\vspace{-8pt}
\caption{\textit{How-learning} explanations for worked examples in fractions and multi-column addition. Of the several explanations, some may be correct (green) or incorrect (red).}
\label{how}
\end{figure}

\textit{How-learning} determines \textit{how} skills apply actions using an abductive process. Prior simulated learners have generally implemented \textit{how-learning} with a search process that composes primitive domain-general prior-knowledge functions (like arithmetic functions and string operations) to reproduce observed actions. This search typically produces multiple candidate compositions that reproduce the worked example, some of which may be incorrect.
Among the candidate compositions that reproduce a worked example, the most parsimonious (having the fewest operations and arguments) is chosen. 
The chosen explanation is generalized by replacing the constants in the grounded composition with variables. For instance, \code{OnesDigit(7+5)} in Figure \ref{how} may be generalized to depend upon two argument variables \code{Arg0=Var(TextField)} and \code{Arg1=Var(TextField)} which match to any TextField type interface elements. 

Prior implementations of \textit{how-learning} have composed primitive functions in an iterative-deepening fashion \cite{matsuda2015teaching}. At each deepening the composition depth is increased by composing primitive functions with prior compositions, for instance, two depth=1 compositions Add(a,b), and Subtract(a,b) could be composed with Divide(a,b) to produce a depth=2 composition Divide(Add(a,b), Subtract(c,d)). This search process explodes combinatorially, so search depths are typically limited to 1 to 3. A method that we have come to call Set Chaining optimizes this search considerably. Set Chaining executes primitive prior knowledge functions in waves, using every combination of unique values from the previous wave as arguments in the next. By keeping a lightweight record of every way each unique value is produced in each wave, compositions can be built by tracing back from the goal value, once it is found. This method cuts down on combinatoric search and is amenable to multithreaded implementations. 

Prior work has also used instructional annotations, like the arguments of the true composition \cite{matsuda2015teaching} and even natural language hints \cite{weitekamp2023simulating}, to guide the explanation process and reduce the amount of combinatorial search.

\subsection{Where-learning}
\begin{figure}[t]
\centering
\includegraphics[width=0.8\columnwidth]{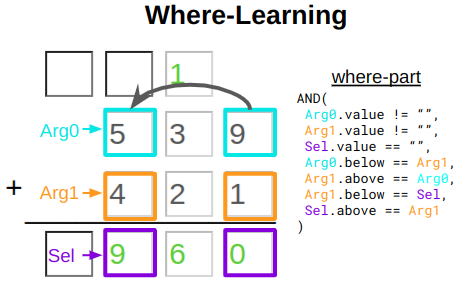}
\vspace{-8pt}
\caption{An example of a \textit{where-part} pattern generalized to act across columns in multi-column addition. }
\label{where}
\end{figure}

\textit{Where-Learning} discovers matching patterns to determine where skills can be applied. While \textit{how-learning} is responsible for producing the operational generalizations within skills, \textit{where-learning} builds spatial generalizations that specify their applicable contexts. In a multi-column addition task (Figure \ref{where}), \textit{where-learning} could generalize a skill for computing the one’s digit of a partial sum so that it applies across columns. The patterns it learns are expressed using argument variables (e.g., \code{Arg0} and \code{Arg1} from the \textit{how-learning} example above), along with a single selection variable (e.g., \code{Sel=Var(TextField)}) that matches the interface element to act upon.

The learned \textit{where-part} pattern consists of a logical statement, expressing necessary conditions and spatial relationships between variables. The pattern produced from an initial worked example is typically highly constrained and may only bind to a limited set of selections and arguments. Subsequent examples can generalize the \textit{where-part} by generalizing or removing the relations that comprise it. \textit{How-learning} has a supporting role in identifying the sets of arguments that \textit{where-learning} generalizes from. \textit{How-learning} attempts to explain each new worked example using existing skills’ \textit{how-part} compositions. If there are any candidate explanations, the one with arguments that would make the minimal change to an existing skill (quantified by a score that measures structure similarity) is used for \textit{where-part} generalization. Otherwise, \textit{how-learning} generates a new skill from the example.

\textit{Where-part} generalizations can also match to neighbors and parents of the selections and arguments, to identify their placement within hierarchical representations. For instance, Li et. al. \cite{li2015integrating} employed representation learning in SimStudent, to learn and match to hierarchies of expressions, terms, coefficients, and variables within algebra equations.


\subsection{When-Learning}

\textit{When-learning} identifies the contexts and order in which skills should be applied. The \textit{when-part} of a skill consists of pre-conditions that define its applicability. \textit{When-learning} is typically implemented using binary classification methods that output symbolic relational expressions. Prior work has used inductive logic programming, decision trees, and incremental concept learning approaches \cite{matsuda2015teaching,maclellan2016apprentice}. In any given state, the learned \textit{when-part} preconditions distinguish whether a particular candidate application of a skill matched by the \textit{where-part} pattern should be applied.


\begin{figure}[t]
\centering
\includegraphics[width=0.9\columnwidth]{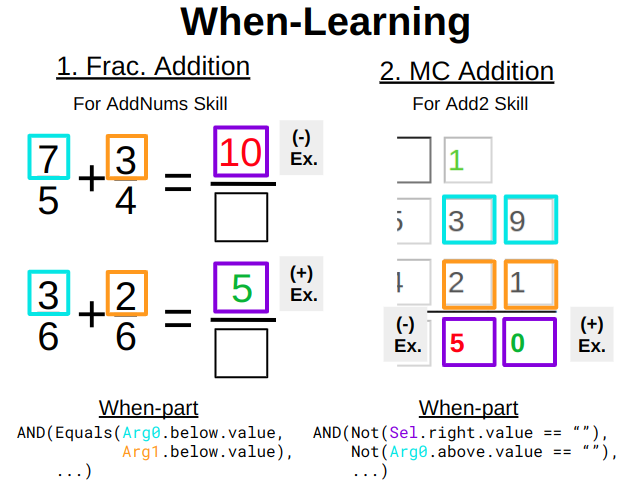} 
\vspace{-8pt}
\caption{Positive and negative examples of an AddNum skill in fraction addition, and an Add2 skill in Multi-Column addition. Partial \textit{when-parts} that accept the positive examples but reject the negative example are shown expressed with relational features generated by relative featurization.}
\label{when}
\end{figure}

\textit{When-learning} uses positive and negative examples of candidate skill applications in particular problem states. \textit{Where-part} processing assists \textit{when-learning} by associating each example triple of (state, action, reward) with a particular skill, selection, and set of arguments. This allows \textit{when-learning} to construct \textit{when-parts} as variablized concepts consisting of relations that express features in the state as they relate to the selection (e.g. Sel) and arguments (e.g. Arg0, Arg1) instead of as they relate to particular interface elements. This enables generalization across spatially distinct instances of the same skill. For instance, the \textit{when-part} in the fraction example in Figure \ref{when} includes a literal relation \code{Equals(Arg0.below.value, Arg1.below.value)} which references the values of the elements below \code{Arg0} and \code{Arg1}. 

Prior work has used FOIL \cite{quinlan1995induction}, an inductive logic programming method, to learn concepts with relational constraints like those above. FOIL searches for and combines new literals like \code{Below(Arg0, A)} expressed in terms of source variables like \code{Arg0}, or invented variables like \code{A}. Searching for such statements can be computationally demanding. In this work we introduce a streamlined means of restating features in the problem state relative to the selection and argument variables. The shortest path through adjacency relationships is found between interface elements using the Bellman-Ford algorithm \cite{bellman1958routing}. Then each feature in the state is relabeled with the shortest path from the selection or arguments (in the dot-notation of Figure \ref{when}). This method of \textit{relative featurization} also allows us to keep the \textit{when-learning} classifier independent of relational feature generation.

\section{Decomposing from RL to DIPL}
Reinforcement Learning (RL) learns policies $\pi(s) \rightarrow a$, directly or indirectly, that map states to actions
to maximize reward over task episodes. RL can maximize overall reward in environments where feedback signals are delayed over states, or even when there are non-deterministic actions. As RL algorithms have evolved and come to rely largely on deep learning, they have shown successes at challenging games \cite{mnih2013playing} and robotics tasks \cite{schulman2017proximal}. They have even become a go-to choice in deterministic procedural domains that require symbolic manipulation, such as theorem proving \cite{kaliszyk2018reinforcement}, and for learning mathematical tasks such as geometry \cite{xiao2023deep}, and early K-12 mathematics like fractions and long arithmetic \cite{poesia2021contrastive}. However, these deep learning approaches typically require at least thousands of examples and generally learn by attempting tasks in specially formatted environments with pre-specified action spaces. 

By comparison, DIPL-based induction is about as data-efficient as human learning \cite{maclellan2016apprentice} and does not require a pre-specified action or state space \cite{maclellan2021EDM}. New ``actions'' are learned through the induction and generalization of skills with \textit{how-} and \textit{where-learning}. \textit{When-learning} then determines in what order and contexts those skills should be applied. By analogy to RL's choice of actions via a global policy $\pi(s) \rightarrow a$ DIPL generates actions with multiple symbolic pieces $When(s,Where(s)) \rightarrow How(Where(s))=a$.


Mathematical and algorithmic complexities notwithstanding, deep RL generally relies on gradient descent to tune its behaviors. In our ablation analysis we will decompose from RL’s 1-mechanism learning to DIPL’s 3-mechanism learning by systematically introducing additional learning mechanisms. Between RL and DIPL a 2-mechanism system employs \textit{how-learning}, but only a single Left-hand-side (LHS) learning mechanism is used to generate correct sets of selections and arguments for producing actions.

\section{Task Domains}

\begin{figure}[t]
\centering
\includegraphics[width=0.75\columnwidth]{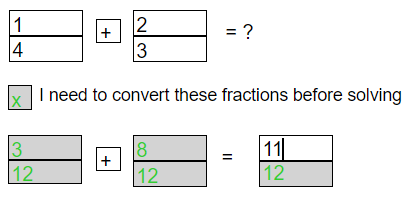} 
\vspace{-8pt}
\caption{Fraction arithmetic tutoring system for teaching multiplying and adding fractions. Learner insert an 'x' to the 'check\_convert' field if the fraction must first be converted.}
\label{frac}
\vspace{-4pt}
\end{figure}

We build on the RL gym environments used by \citeA{maclellan2021EDM} for two different ITS domains: (1) A fraction arithmetic tutor (Figure \ref{frac}) that randomly selects among: adding same denominator fractions, different denominator fractions, and multiplying fractions and (2) a multi-column arithmetic tutor that teaches 3-digit addition (Figures \ref{where},\ref{when}). In both domains, agents can request worked examples (demos), and receive immediate reward signals (1 for correct, -1 for incorrect) on attempted actions. Since action spaces consist only of, checking boxes, placing numbers in fields, or pressing the `done' button, there are finite primitive actions for an RL system to select from.


In the fractions tutor, agents must perform the correct fraction arithmetic procedure step-by-step (multiplying, adding, or converting then adding) based on the two starting fractions and the operator. In the RL-Gym wrapper for this environment, the agent is able to fill in each of 6 number fields with the numbers 1-450, fill in the `check\_convert' field with an ``x'' or press the done button, for a total of 2,702 unique actions. The multi-column addition domain (see Figures \ref{where}, \ref{when}) has 7 fields that can be filled with the digits 0-9, plus a done button for a total of 71 unique actions. In this domain, the agent must compute the sum of two 3-digit numbers by computing each partial sum in right-to-left order by placing the ones digit and then carrying the tens-digit when necessary. In both domains, the state is encoded into a vector with 0.0 or 1.0 representing whether each element is present using one-hot encoding (size 2,000 in fractions and 240 in multi-column addition). The one-hot encoding maps each unique interface element-attribute pair to a slot in the state vector. 

The non-RL-based agents instead experience the state in its original object-based representation, where each object has a unique identifier, type, position, shape, and value. Additionally, no predefined action space is provided to these agents. Instead, they learn how to produce actions by applying \textit{how-learning} to the on-demand worked example demos. These agents are instantiated with the primitive domain-general prior knowledge functions necessary to compose \textit{how-parts} for each task: \code{Add(a,b)} and \code{Multiply(a,b)} for fractions and \code{OnesDigit(a)}, \code{TensDigit(a)}, \code{Add3(a,b,c)}, \code{Add(a,b)} for multi-column addition. In multi-column addition, extraneous \textit{how-learning} explanations are common, so we aid \textit{how-learning} by annotating each demo with its arguments. We do not provide these annotations for fractions. In the fractions domain, we provided a single feature function \code{Equals(a,b)}, enabling the agent to identify equal values, which is necessary for learning to check for equal denominators. 

\section{Ablation Analysis}
We apply two RL approaches: an off-policy Deep-Q-Network (DQN) model \cite{mnih2015dqn} and an on-policy Proximal Policy Optimization (PPO) model \cite{schulman2017proximal}. PPO has become a popular RL approach for its relative stability, data-efficiency, and consistent convergence without hyperparameter tuning. We additionally train agents with or without automatically provided worked examples. In the latter case (indicated by “+Demos”), each incorrect action is followed by training on the current step's demo worked example. This mimics the capability of DIPL-based simulated learners to request demos when no next action can be produced. Unfortunately, this method only works with the DQN models, as there is no simple method for training on-policy methods like PPO with actions not produced by its current policy. All models were implemented using OpenAI’s stable baselines library and were trained for 500,000 timesteps. For a symbolic comparison, we additionally train a decision tree using the “+Demos” training modality. 

Our 2-mechanism model uses a Set Chaining \textit{how-learning} mechanism to learn individual skills, but only a single Left-Hand-Side (LHS) learning mechanism that predicts where and when those skills should be applied. The LHS-learning uses a decision tree as a multi-class classifier that predicts the selection and arguments for the correct next action from features of the problem state.

Finally, we utilize a DIPL-base agent via our re-implementation of the Apprentice Learner (AL) Architecture. Set Chaining is used for \textit{how-learning}. \textit{Where-learning} uses a simple implementation that only recalls sets of selections and arguments from past examples. Finally, \textit{when-learning} is achieved with a decision tree. We train both with and without relative featurization to illustrate the effects of utilizing \textit{where-part} processing in \textit{when-learning}.

\subsection{Results}
\begin{table}[]
\begin{tabular}{ll|l|l|}
\cline{3-4}
 &  & Fractions & MC Addition \\ \hline
\multicolumn{1}{|l|}{\multirow{3}{*}{1-mech}} & PPO & Not Converge & 30,642 \\ \cline{2-4} 
\multicolumn{1}{|l|}{} & DQN+Demos & 11,315 & 9,496 \\ \cline{2-4} 
\multicolumn{1}{|l|}{} & DT+Demos & 1,944 & 7,816 \\ \hline
\multicolumn{1}{|l|}{2-mech} & How+LHS & 17 & 270 \\ \hline
\multicolumn{1}{|l|}{\multirow{2}{*}{3-mech}} & \begin{tabular}[c]{@{}l@{}}DIPL \\ (no rel. feat.)\end{tabular} & 33 & 38 \\ \cline{2-4} 
\multicolumn{1}{|l|}{} & DIPL & 20 & 19 \\ \hline
\multicolumn{1}{|l|}{} & Human Data & $\sim$9-14 & N/A \\ \hline
\end{tabular}
\caption{Number of problems before $<10\%$ average error.}
\end{table}

\begin{figure}[t]
\centering
\includegraphics[width=0.99\columnwidth]{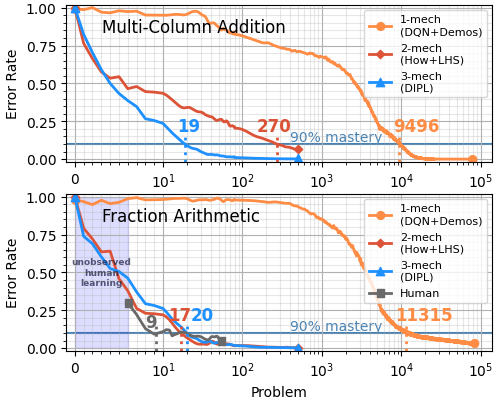} 
\vspace{-8pt}
\caption{Log-scale x-axis learning curves, for DQN-Demos, How+LHS, DIPL annotated with 10\% error intercept. Fraction domain includes human curves (grey) offset to account for unobserved learning opportunities. }
\label{curves}
\end{figure}

Our results are summarized in Table 1 and select learning curves are shown in Figure \ref{curves}. The DQN models converged only in the "+Demos" condition. PPO successfully converged only in multi-column addition. Among the RL methods, the “DQN + demos” approach achieved the best data efficiency up to the $<$10\% error mastery point requiring about 10,000 problems in each task. Decision trees were more data-efficient than the RL methods, but still required 1,944 problems in fractions and 7,816 problems in multi-column addition. The 2-mechanism model showed a dramatic improvement in data efficiency, taking just 17 problems to master fractions but 270 for multi-column addition. The 3-mechanism DIPL model by contrast mastered both in 20 problems or less. Turning off relative featurization resulted in a 13-19 problem deterioration of data efficiency. We additionally include human data collected by \citeA{patel2016block} using the fractions ITS. We shift this data by 6 problems to align the initial 30\% error rate exhibited by the most efficient SL, leading to an adjusted mastery intercept of around 9-14 problems. The DIPL agents and humans show similar initial learning rates, but the DIPL agents improve more rapidly beyond the mastery threshold up to less than 1\% error after about 130 problems.

\subsection{Discussion}
\subsubsection{Data-Efficiency} Our 1-mechanism models cover only a small number of RL training approaches yet are fairly representative of the data-efficiency of RL. One contribution to inefficiency is that the RL action models consist of picking numbers instead of computing them (similar to how LLMs approach math). However, in similar experiments in which RL agents were given domain-specific primitive actions equivalent to what an SL would induce through \textit{how-learning}, training still required thousands of episodes \cite{maclellan2021EDM}. 

The decision tree's relatively better data efficiency demonstrates an advantage of symbolic versus sub-symbolic learning. However, learning decomposition proved to be more essential to achieving human-like data efficiency. In the 2-mechanism model, \textit{how-learning} helped produce near human-like efficiency for fractions, but in multi-column tutor a further decomposition into \textit{where-} and \textit{when-learning} was essential.  \textit{Where-learning} lets \textit{when-learning} spatially generalize across multiple uses of the same skill (e.g., across columns).


\subsubsection{Further Decomposition}
We may consider if yet more decomposition could produce greater data efficiency. For instance, HTN induction in VanLehn's \citeyear{vanlehn1990mind} Sierra, and the \textit{process-learning} mechanism in AI2T may simplify the role of \textit{when-learning}. When situated within an HTN, skills can be ordered explicitly within methods that dictate the steps for accomplishing higher-level tasks. This may reduce the role of \textit{when-learning} so that it only needs to learn simple preconditions that gate the applicability of methods instead of also controlling the order in which primitive actions are applied. Generalizing this approach to work beyond the limited domains on which it has been applied may be a path toward yet greater data efficiency than we achieved here.

\section{Conclusion}
The rise of data-driven machine learning and LLMs has inspired a fixation on replicating human performance while overlooking the vast gap in data efficiency between data-driven machine learning and human learning, which is orders of magnitude more data efficient. This work demonstrates how cooperation between specialized learning mechanisms is a potential path to bridging that gap---enabling learning from tens of examples instead of tens of thousands.


\bibliographystyle{apacite}

\setlength{\bibleftmargin}{.125in}
\setlength{\bibindent}{-\bibleftmargin}

\bibliography{references}

\begin{thebibliography}{}

\bibitem [\protect \citeauthoryear {%
Anderson%
\ \BBA {} Schunn%
}{%
Anderson%
\ \BBA {} Schunn%
}{%
{\protect \APACyear {2013}}%
}]{%
anderson2013implications}
\APACinsertmetastar {%
anderson2013implications}%
\begin{APACrefauthors}%
Anderson, J\BPBI R.%
\BCBT {}\ \BBA {} Schunn, C\BPBI D.%
\end{APACrefauthors}%
\unskip\
\newblock
\APACrefYearMonthDay{2013}{}{}.
\newblock
{\BBOQ}\APACrefatitle {Implications of the ACT-R learning theory: No magic bullets} {Implications of the act-r learning theory: No magic bullets}.{\BBCQ}
\newblock
\BIn{} \APACrefbtitle {Advances in instructional Psychology, Volume 5} {Advances in instructional psychology, volume 5}\ (\BPGS\ 1--33).
\newblock
\APACaddressPublisher{}{Routledge}.
\PrintBackRefs{\CurrentBib}

\bibitem [\protect \citeauthoryear {%
Bellman%
}{%
Bellman%
}{%
{\protect \APACyear {1958}}%
}]{%
bellman1958routing}
\APACinsertmetastar {%
bellman1958routing}%
\begin{APACrefauthors}%
Bellman, R.%
\end{APACrefauthors}%
\unskip\
\newblock
\APACrefYearMonthDay{1958}{}{}.
\newblock
{\BBOQ}\APACrefatitle {On a routing problem} {On a routing problem}.{\BBCQ}
\newblock
\APACjournalVolNumPages{Quarterly of applied mathematics}{16}{1}{87--90}.
\PrintBackRefs{\CurrentBib}

\bibitem [\protect \citeauthoryear {%
Bubeck%
\ \protect \BOthers {.}}{%
Bubeck%
\ \protect \BOthers {.}}{%
{\protect \APACyear {2023}}%
}]{%
bubeck2023sparks}
\APACinsertmetastar {%
bubeck2023sparks}%
\begin{APACrefauthors}%
Bubeck, S.%
, Chandrasekaran, V.%
, Eldan, R.%
, Gehrke, J.%
, Horvitz, E.%
, Kamar, E.%
\BDBL {}others%
\end{APACrefauthors}%
\unskip\
\newblock
\APACrefYearMonthDay{2023}{}{}.
\newblock
{\BBOQ}\APACrefatitle {Sparks of artificial general intelligence: Early experiments with gpt-4} {Sparks of artificial general intelligence: Early experiments with gpt-4}.{\BBCQ}
\newblock
\APACjournalVolNumPages{arXiv preprint arXiv:2303.12712}{}{}{}.
\PrintBackRefs{\CurrentBib}

\bibitem [\protect \citeauthoryear {%
Cobbe%
\ \protect \BOthers {.}}{%
Cobbe%
\ \protect \BOthers {.}}{%
{\protect \APACyear {2021}}%
}]{%
cobbe2021training}
\APACinsertmetastar {%
cobbe2021training}%
\begin{APACrefauthors}%
Cobbe, K.%
, Kosaraju, V.%
, Bavarian, M.%
, Chen, M.%
, Jun, H.%
, Kaiser, L.%
\BDBL {}others%
\end{APACrefauthors}%
\unskip\
\newblock
\APACrefYearMonthDay{2021}{}{}.
\newblock
{\BBOQ}\APACrefatitle {Training verifiers to solve math word problems, 2021} {Training verifiers to solve math word problems, 2021}.{\BBCQ}
\newblock
\APACjournalVolNumPages{URL https://arxiv. org/abs/2110.14168}{}{}{}.
\PrintBackRefs{\CurrentBib}

\bibitem [\protect \citeauthoryear {%
Erol%
, Hendler%
\BCBL {}\ \BBA {} Nau%
}{%
Erol%
\ \protect \BOthers {.}}{%
{\protect \APACyear {1994}}%
}]{%
erol1994semantics}
\APACinsertmetastar {%
erol1994semantics}%
\begin{APACrefauthors}%
Erol, K.%
, Hendler, J\BPBI A.%
\BCBL {}\ \BBA {} Nau, D\BPBI S.%
\end{APACrefauthors}%
\unskip\
\newblock
\APACrefYear{1994}.
\newblock
\APACrefbtitle {Semantics for hierarchical task-network planning} {Semantics for hierarchical task-network planning}.
\newblock
\APACaddressPublisher{}{Citeseer}.
\PrintBackRefs{\CurrentBib}

\bibitem [\protect \citeauthoryear {%
Hendrycks%
\ \protect \BOthers {.}}{%
Hendrycks%
\ \protect \BOthers {.}}{%
{\protect \APACyear {2021}}%
}]{%
hendrycks2021measuring}
\APACinsertmetastar {%
hendrycks2021measuring}%
\begin{APACrefauthors}%
Hendrycks, D.%
, Burns, C.%
, Kadavath, S.%
, Arora, A.%
, Basart, S.%
, Tang, E.%
\BDBL {}Steinhardt, J.%
\end{APACrefauthors}%
\unskip\
\newblock
\APACrefYearMonthDay{2021}{}{}.
\newblock
{\BBOQ}\APACrefatitle {Measuring mathematical problem solving with the math dataset} {Measuring mathematical problem solving with the math dataset}.{\BBCQ}
\newblock
\APACjournalVolNumPages{arXiv preprint arXiv:2103.03874}{}{}{}.
\PrintBackRefs{\CurrentBib}

\bibitem [\protect \citeauthoryear {%
Kaliszyk%
, Urban%
, Michalewski%
\BCBL {}\ \BBA {} Ol{\v{s}}{\'a}k%
}{%
Kaliszyk%
\ \protect \BOthers {.}}{%
{\protect \APACyear {2018}}%
}]{%
kaliszyk2018reinforcement}
\APACinsertmetastar {%
kaliszyk2018reinforcement}%
\begin{APACrefauthors}%
Kaliszyk, C.%
, Urban, J.%
, Michalewski, H.%
\BCBL {}\ \BBA {} Ol{\v{s}}{\'a}k, M.%
\end{APACrefauthors}%
\unskip\
\newblock
\APACrefYearMonthDay{2018}{}{}.
\newblock
{\BBOQ}\APACrefatitle {Reinforcement learning of theorem proving} {Reinforcement learning of theorem proving}.{\BBCQ}
\newblock
\APACjournalVolNumPages{Advances in Neural Information Processing Systems}{31}{}{}.
\PrintBackRefs{\CurrentBib}

\bibitem [\protect \citeauthoryear {%
K{\"a}ser%
\ \BBA {} Alexandron%
}{%
K{\"a}ser%
\ \BBA {} Alexandron%
}{%
{\protect \APACyear {2024}}%
}]{%
kaser2024simulated}
\APACinsertmetastar {%
kaser2024simulated}%
\begin{APACrefauthors}%
K{\"a}ser, T.%
\BCBT {}\ \BBA {} Alexandron, G.%
\end{APACrefauthors}%
\unskip\
\newblock
\APACrefYearMonthDay{2024}{}{}.
\newblock
{\BBOQ}\APACrefatitle {Simulated learners in educational technology: A systematic literature review and a turing-like test} {Simulated learners in educational technology: A systematic literature review and a turing-like test}.{\BBCQ}
\newblock
\APACjournalVolNumPages{International Journal of Artificial Intelligence in Education}{34}{2}{545--585}.
\PrintBackRefs{\CurrentBib}

\bibitem [\protect \citeauthoryear {%
Koedinger%
\ \protect \BOthers {.}}{%
Koedinger%
\ \protect \BOthers {.}}{%
{\protect \APACyear {2010}}%
}]{%
koedinger2010data}
\APACinsertmetastar {%
koedinger2010data}%
\begin{APACrefauthors}%
Koedinger, K\BPBI R.%
, Baker, R\BPBI S.%
, Cunningham, K.%
, Skogsholm, A.%
, Leber, B.%
\BCBL {}\ \BBA {} Stamper, J.%
\end{APACrefauthors}%
\unskip\
\newblock
\APACrefYearMonthDay{2010}{}{}.
\newblock
{\BBOQ}\APACrefatitle {A data repository for the EDM community: The PSLC DataShop} {A data repository for the edm community: The pslc datashop}.{\BBCQ}
\newblock
\APACjournalVolNumPages{Handbook of educational data mining}{43}{}{43--56}.
\PrintBackRefs{\CurrentBib}

\bibitem [\protect \citeauthoryear {%
Koedinger%
, Carvalho%
, Liu%
\BCBL {}\ \BBA {} McLaughlin%
}{%
Koedinger%
\ \protect \BOthers {.}}{%
{\protect \APACyear {2023}}%
}]{%
koedinger2023astonishing}
\APACinsertmetastar {%
koedinger2023astonishing}%
\begin{APACrefauthors}%
Koedinger, K\BPBI R.%
, Carvalho, P\BPBI F.%
, Liu, R.%
\BCBL {}\ \BBA {} McLaughlin, E\BPBI A.%
\end{APACrefauthors}%
\unskip\
\newblock
\APACrefYearMonthDay{2023}{}{}.
\newblock
{\BBOQ}\APACrefatitle {An astonishing regularity in student learning rate} {An astonishing regularity in student learning rate}.{\BBCQ}
\newblock
\APACjournalVolNumPages{Proceedings of the National Academy of Sciences}{120}{13}{e2221311120}.
\PrintBackRefs{\CurrentBib}

\bibitem [\protect \citeauthoryear {%
Koedinger%
, Corbett%
\BCBL {}\ \BBA {} Perfetti%
}{%
Koedinger%
\ \protect \BOthers {.}}{%
{\protect \APACyear {2012}}%
}]{%
koedinger2012knowledge}
\APACinsertmetastar {%
koedinger2012knowledge}%
\begin{APACrefauthors}%
Koedinger, K\BPBI R.%
, Corbett, A\BPBI T.%
\BCBL {}\ \BBA {} Perfetti, C.%
\end{APACrefauthors}%
\unskip\
\newblock
\APACrefYearMonthDay{2012}{}{}.
\newblock
{\BBOQ}\APACrefatitle {The Knowledge-Learning-Instruction framework: Bridging the science-practice chasm to enhance robust student learning} {The knowledge-learning-instruction framework: Bridging the science-practice chasm to enhance robust student learning}.{\BBCQ}
\newblock
\APACjournalVolNumPages{Cognitive science}{36}{5}{757--798}.
\PrintBackRefs{\CurrentBib}

\bibitem [\protect \citeauthoryear {%
Konda%
\ \BBA {} Tsitsiklis%
}{%
Konda%
\ \BBA {} Tsitsiklis%
}{%
{\protect \APACyear {1999}}%
}]{%
konda1999actor}
\APACinsertmetastar {%
konda1999actor}%
\begin{APACrefauthors}%
Konda, V.%
\BCBT {}\ \BBA {} Tsitsiklis, J.%
\end{APACrefauthors}%
\unskip\
\newblock
\APACrefYearMonthDay{1999}{}{}.
\newblock
{\BBOQ}\APACrefatitle {Actor-critic algorithms} {Actor-critic algorithms}.{\BBCQ}
\newblock
\APACjournalVolNumPages{Advances in neural information processing systems}{12}{}{}.
\PrintBackRefs{\CurrentBib}

\bibitem [\protect \citeauthoryear {%
Laird%
}{%
Laird%
}{%
{\protect \APACyear {2019}}%
}]{%
laird2019soar}
\APACinsertmetastar {%
laird2019soar}%
\begin{APACrefauthors}%
Laird, J\BPBI E.%
\end{APACrefauthors}%
\unskip\
\newblock
\APACrefYear{2019}.
\newblock
\APACrefbtitle {The Soar cognitive architecture} {The soar cognitive architecture}.
\newblock
\APACaddressPublisher{}{MIT press}.
\PrintBackRefs{\CurrentBib}

\bibitem [\protect \citeauthoryear {%
Laird%
, Lebiere%
\BCBL {}\ \BBA {} Rosenbloom%
}{%
Laird%
\ \protect \BOthers {.}}{%
{\protect \APACyear {2017}}%
}]{%
laird2017standard}
\APACinsertmetastar {%
laird2017standard}%
\begin{APACrefauthors}%
Laird, J\BPBI E.%
, Lebiere, C.%
\BCBL {}\ \BBA {} Rosenbloom, P\BPBI S.%
\end{APACrefauthors}%
\unskip\
\newblock
\APACrefYearMonthDay{2017}{}{}.
\newblock
{\BBOQ}\APACrefatitle {A standard model of the mind: Toward a common computational framework across artificial intelligence, cognitive science, neuroscience, and robotics} {A standard model of the mind: Toward a common computational framework across artificial intelligence, cognitive science, neuroscience, and robotics}.{\BBCQ}
\newblock
\APACjournalVolNumPages{Ai Magazine}{38}{4}{13--26}.
\PrintBackRefs{\CurrentBib}

\bibitem [\protect \citeauthoryear {%
Li%
, Matsuda%
, Cohen%
\BCBL {}\ \BBA {} Koedinger%
}{%
Li%
\ \protect \BOthers {.}}{%
{\protect \APACyear {2015}}%
}]{%
li2015integrating}
\APACinsertmetastar {%
li2015integrating}%
\begin{APACrefauthors}%
Li, N.%
, Matsuda, N.%
, Cohen, W\BPBI W.%
\BCBL {}\ \BBA {} Koedinger, K\BPBI R.%
\end{APACrefauthors}%
\unskip\
\newblock
\APACrefYearMonthDay{2015}{}{}.
\newblock
{\BBOQ}\APACrefatitle {Integrating representation learning and skill learning in a human-like intelligent agent} {Integrating representation learning and skill learning in a human-like intelligent agent}.{\BBCQ}
\newblock
\APACjournalVolNumPages{Artificial Intelligence}{219}{}{67--91}.
\PrintBackRefs{\CurrentBib}

\bibitem [\protect \citeauthoryear {%
Lightman%
\ \protect \BOthers {.}}{%
Lightman%
\ \protect \BOthers {.}}{%
{\protect \APACyear {2023}}%
}]{%
lightman2023let}
\APACinsertmetastar {%
lightman2023let}%
\begin{APACrefauthors}%
Lightman, H.%
, Kosaraju, V.%
, Burda, Y.%
, Edwards, H.%
, Baker, B.%
, Lee, T.%
\BDBL {}Cobbe, K.%
\end{APACrefauthors}%
\unskip\
\newblock
\APACrefYearMonthDay{2023}{}{}.
\newblock
{\BBOQ}\APACrefatitle {Let's Verify Step by Step} {Let's verify step by step}.{\BBCQ}
\newblock
\APACjournalVolNumPages{arXiv preprint arXiv:2305.20050}{}{}{}.
\PrintBackRefs{\CurrentBib}

\bibitem [\protect \citeauthoryear {%
MacLellan%
\ \BBA {} Gupta%
}{%
MacLellan%
\ \BBA {} Gupta%
}{%
{\protect \APACyear {2021}}%
}]{%
maclellan2021EDM}
\APACinsertmetastar {%
maclellan2021EDM}%
\begin{APACrefauthors}%
MacLellan, C\BPBI J.%
\BCBT {}\ \BBA {} Gupta, A.%
\end{APACrefauthors}%
\unskip\
\newblock
\APACrefYearMonthDay{2021}{}{}.
\newblock
{\BBOQ}\APACrefatitle {Learning Expert Models for Educationally Relevant Tasks Using Reinforcement Learning.} {Learning expert models for educationally relevant tasks using reinforcement learning.}{\BBCQ}
\newblock
\APACjournalVolNumPages{International Educational Data Mining Society}{}{}{}.
\PrintBackRefs{\CurrentBib}

\bibitem [\protect \citeauthoryear {%
Maclellan%
, Harpstead%
, Patel%
\BCBL {}\ \BBA {} Koedinger%
}{%
Maclellan%
\ \protect \BOthers {.}}{%
{\protect \APACyear {2016}}%
}]{%
maclellan2016apprentice}
\APACinsertmetastar {%
maclellan2016apprentice}%
\begin{APACrefauthors}%
Maclellan, C\BPBI J.%
, Harpstead, E.%
, Patel, R.%
\BCBL {}\ \BBA {} Koedinger, K\BPBI R.%
\end{APACrefauthors}%
\unskip\
\newblock
\APACrefYearMonthDay{2016}{}{}.
\newblock
{\BBOQ}\APACrefatitle {The Apprentice Learner Architecture: Closing the Loop between Learning Theory and Educational Data.} {The apprentice learner architecture: Closing the loop between learning theory and educational data.}{\BBCQ}
\newblock
\APACjournalVolNumPages{International Educational Data Mining Society}{}{}{}.
\PrintBackRefs{\CurrentBib}

\bibitem [\protect \citeauthoryear {%
MacLellan%
\ \BBA {} Koedinger%
}{%
MacLellan%
\ \BBA {} Koedinger%
}{%
{\protect \APACyear {2020}}%
}]{%
maclellan2020domain}
\APACinsertmetastar {%
maclellan2020domain}%
\begin{APACrefauthors}%
MacLellan, C\BPBI J.%
\BCBT {}\ \BBA {} Koedinger, K\BPBI R.%
\end{APACrefauthors}%
\unskip\
\newblock
\APACrefYearMonthDay{2020}{}{}.
\newblock
{\BBOQ}\APACrefatitle {Domain-General Tutor Authoring with Apprentice Learner Models} {Domain-general tutor authoring with apprentice learner models}.{\BBCQ}
\newblock
\APACjournalVolNumPages{International Journal of Artificial Intelligence in Education}{}{}{1--42}.
\PrintBackRefs{\CurrentBib}

\bibitem [\protect \citeauthoryear {%
Manhaeve%
, Dumancic%
, Kimmig%
, Demeester%
\BCBL {}\ \BBA {} De~Raedt%
}{%
Manhaeve%
\ \protect \BOthers {.}}{%
{\protect \APACyear {2018}}%
}]{%
manhaeve2018deepproblog}
\APACinsertmetastar {%
manhaeve2018deepproblog}%
\begin{APACrefauthors}%
Manhaeve, R.%
, Dumancic, S.%
, Kimmig, A.%
, Demeester, T.%
\BCBL {}\ \BBA {} De~Raedt, L.%
\end{APACrefauthors}%
\unskip\
\newblock
\APACrefYearMonthDay{2018}{}{}.
\newblock
{\BBOQ}\APACrefatitle {Deepproblog: Neural probabilistic logic programming} {Deepproblog: Neural probabilistic logic programming}.{\BBCQ}
\newblock
\APACjournalVolNumPages{Advances in neural information processing systems}{31}{}{}.
\PrintBackRefs{\CurrentBib}

\bibitem [\protect \citeauthoryear {%
Matsuda%
, Cohen%
\BCBL {}\ \BBA {} Koedinger%
}{%
Matsuda%
\ \protect \BOthers {.}}{%
{\protect \APACyear {2015}}%
}]{%
matsuda2015teaching}
\APACinsertmetastar {%
matsuda2015teaching}%
\begin{APACrefauthors}%
Matsuda, N.%
, Cohen, W\BPBI W.%
\BCBL {}\ \BBA {} Koedinger, K\BPBI R.%
\end{APACrefauthors}%
\unskip\
\newblock
\APACrefYearMonthDay{2015}{}{}.
\newblock
{\BBOQ}\APACrefatitle {Teaching the Teacher: Tutoring SimStudent Leads to More Effective Cognitive Tutor Authoring} {Teaching the teacher: Tutoring simstudent leads to more effective cognitive tutor authoring}.{\BBCQ}
\newblock
\APACjournalVolNumPages{International Journal of Artificial Intelligence in Education}{25}{1}{1--34}.
\PrintBackRefs{\CurrentBib}

\bibitem [\protect \citeauthoryear {%
McCloskey%
\ \BBA {} Cohen%
}{%
McCloskey%
\ \BBA {} Cohen%
}{%
{\protect \APACyear {1989}}%
}]{%
mccloskey1989catastrophic}
\APACinsertmetastar {%
mccloskey1989catastrophic}%
\begin{APACrefauthors}%
McCloskey, M.%
\BCBT {}\ \BBA {} Cohen, N\BPBI J.%
\end{APACrefauthors}%
\unskip\
\newblock
\APACrefYearMonthDay{1989}{}{}.
\newblock
{\BBOQ}\APACrefatitle {Catastrophic interference in connectionist networks: The sequential learning problem} {Catastrophic interference in connectionist networks: The sequential learning problem}.{\BBCQ}
\newblock
\BIn{} \APACrefbtitle {Psychology of learning and motivation} {Psychology of learning and motivation}\ (\BVOL~24, \BPGS\ 109--165).
\newblock
\APACaddressPublisher{}{Elsevier}.
\PrintBackRefs{\CurrentBib}

\bibitem [\protect \citeauthoryear {%
Mirzadeh%
\ \protect \BOthers {.}}{%
Mirzadeh%
\ \protect \BOthers {.}}{%
{\protect \APACyear {2024}}%
}]{%
mirzadeh2024gsm}
\APACinsertmetastar {%
mirzadeh2024gsm}%
\begin{APACrefauthors}%
Mirzadeh, I.%
, Alizadeh, K.%
, Shahrokhi, H.%
, Tuzel, O.%
, Bengio, S.%
\BCBL {}\ \BBA {} Farajtabar, M.%
\end{APACrefauthors}%
\unskip\
\newblock
\APACrefYearMonthDay{2024}{}{}.
\newblock
{\BBOQ}\APACrefatitle {Gsm-symbolic: Understanding the limitations of mathematical reasoning in large language models} {Gsm-symbolic: Understanding the limitations of mathematical reasoning in large language models}.{\BBCQ}
\newblock
\APACjournalVolNumPages{arXiv preprint arXiv:2410.05229}{}{}{}.
\PrintBackRefs{\CurrentBib}

\bibitem [\protect \citeauthoryear {%
Mnih%
\ \protect \BOthers {.}}{%
Mnih%
\ \protect \BOthers {.}}{%
{\protect \APACyear {2013}}%
}]{%
mnih2013playing}
\APACinsertmetastar {%
mnih2013playing}%
\begin{APACrefauthors}%
Mnih, V.%
, Kavukcuoglu, K.%
, Silver, D.%
, Graves, A.%
, Antonoglou, I.%
, Wierstra, D.%
\BCBL {}\ \BBA {} Riedmiller, M.%
\end{APACrefauthors}%
\unskip\
\newblock
\APACrefYearMonthDay{2013}{}{}.
\newblock
{\BBOQ}\APACrefatitle {Playing atari with deep reinforcement learning} {Playing atari with deep reinforcement learning}.{\BBCQ}
\newblock
\APACjournalVolNumPages{arXiv preprint arXiv:1312.5602}{}{}{}.
\PrintBackRefs{\CurrentBib}

\bibitem [\protect \citeauthoryear {%
Mnih%
\ \protect \BOthers {.}}{%
Mnih%
\ \protect \BOthers {.}}{%
{\protect \APACyear {2015}}%
}]{%
mnih2015dqn}
\APACinsertmetastar {%
mnih2015dqn}%
\begin{APACrefauthors}%
Mnih, V.%
, Kavukcuoglu, K.%
, Silver, D.%
, Rusu, A\BPBI A.%
, Veness, J.%
, Bellemare, M\BPBI G.%
\BDBL {}others%
\end{APACrefauthors}%
\unskip\
\newblock
\APACrefYearMonthDay{2015}{}{}.
\newblock
{\BBOQ}\APACrefatitle {Human-level control through deep reinforcement learning} {Human-level control through deep reinforcement learning}.{\BBCQ}
\newblock
\APACjournalVolNumPages{nature}{518}{7540}{529--533}.
\PrintBackRefs{\CurrentBib}

\bibitem [\protect \citeauthoryear {%
Neves%
}{%
Neves%
}{%
{\protect \APACyear {1985}}%
}]{%
neves1985learning}
\APACinsertmetastar {%
neves1985learning}%
\begin{APACrefauthors}%
Neves, D\BPBI M.%
\end{APACrefauthors}%
\unskip\
\newblock
\APACrefYearMonthDay{1985}{}{}.
\newblock
{\BBOQ}\APACrefatitle {Learning Procedures from Examples and by Doing.} {Learning procedures from examples and by doing.}{\BBCQ}
\newblock
\BIn{} \APACrefbtitle {IJCAI} {Ijcai}\ (\BPGS\ 624--630).
\PrintBackRefs{\CurrentBib}

\bibitem [\protect \citeauthoryear {%
Patel%
, Liu%
\BCBL {}\ \BBA {} Koedinger%
}{%
Patel%
\ \protect \BOthers {.}}{%
{\protect \APACyear {2016}}%
}]{%
patel2016block}
\APACinsertmetastar {%
patel2016block}%
\begin{APACrefauthors}%
Patel, R.%
, Liu, R.%
\BCBL {}\ \BBA {} Koedinger, K\BPBI R.%
\end{APACrefauthors}%
\unskip\
\newblock
\APACrefYearMonthDay{2016}{}{}.
\newblock
{\BBOQ}\APACrefatitle {When to Block versus Interleave Practice? Evidence Against Teaching Fraction Addition before Fraction Multiplication.} {When to block versus interleave practice? evidence against teaching fraction addition before fraction multiplication.}{\BBCQ}
\newblock
\BIn{} \APACrefbtitle {CogSci.} {Cogsci.}
\PrintBackRefs{\CurrentBib}

\bibitem [\protect \citeauthoryear {%
Poesia%
, Dong%
\BCBL {}\ \BBA {} Goodman%
}{%
Poesia%
\ \protect \BOthers {.}}{%
{\protect \APACyear {2021}}%
}]{%
poesia2021contrastive}
\APACinsertmetastar {%
poesia2021contrastive}%
\begin{APACrefauthors}%
Poesia, G.%
, Dong, W.%
\BCBL {}\ \BBA {} Goodman, N.%
\end{APACrefauthors}%
\unskip\
\newblock
\APACrefYearMonthDay{2021}{}{}.
\newblock
{\BBOQ}\APACrefatitle {Contrastive reinforcement learning of symbolic reasoning domains} {Contrastive reinforcement learning of symbolic reasoning domains}.{\BBCQ}
\newblock
\APACjournalVolNumPages{Advances in neural information processing systems}{34}{}{15946--15956}.
\PrintBackRefs{\CurrentBib}

\bibitem [\protect \citeauthoryear {%
Quinlan%
\ \BBA {} Cameron-Jones%
}{%
Quinlan%
\ \BBA {} Cameron-Jones%
}{%
{\protect \APACyear {1995}}%
}]{%
quinlan1995induction}
\APACinsertmetastar {%
quinlan1995induction}%
\begin{APACrefauthors}%
Quinlan, J\BPBI R.%
\BCBT {}\ \BBA {} Cameron-Jones, R\BPBI M.%
\end{APACrefauthors}%
\unskip\
\newblock
\APACrefYearMonthDay{1995}{}{}.
\newblock
{\BBOQ}\APACrefatitle {Induction of logic programs: FOIL and related systems} {Induction of logic programs: Foil and related systems}.{\BBCQ}
\newblock
\APACjournalVolNumPages{New Generation Computing}{13}{}{287--312}.
\PrintBackRefs{\CurrentBib}

\bibitem [\protect \citeauthoryear {%
Ritter%
, Tehranchi%
\BCBL {}\ \BBA {} Oury%
}{%
Ritter%
\ \protect \BOthers {.}}{%
{\protect \APACyear {2019}}%
}]{%
ritter2019act}
\APACinsertmetastar {%
ritter2019act}%
\begin{APACrefauthors}%
Ritter, F\BPBI E.%
, Tehranchi, F.%
\BCBL {}\ \BBA {} Oury, J\BPBI D.%
\end{APACrefauthors}%
\unskip\
\newblock
\APACrefYearMonthDay{2019}{}{}.
\newblock
{\BBOQ}\APACrefatitle {ACT-R: A cognitive architecture for modeling cognition} {Act-r: A cognitive architecture for modeling cognition}.{\BBCQ}
\newblock
\APACjournalVolNumPages{Wiley Interdisciplinary Reviews: Cognitive Science}{10}{3}{e1488}.
\PrintBackRefs{\CurrentBib}

\bibitem [\protect \citeauthoryear {%
Schulman%
, Wolski%
, Dhariwal%
, Radford%
\BCBL {}\ \BBA {} Klimov%
}{%
Schulman%
\ \protect \BOthers {.}}{%
{\protect \APACyear {2017}}%
}]{%
schulman2017proximal}
\APACinsertmetastar {%
schulman2017proximal}%
\begin{APACrefauthors}%
Schulman, J.%
, Wolski, F.%
, Dhariwal, P.%
, Radford, A.%
\BCBL {}\ \BBA {} Klimov, O.%
\end{APACrefauthors}%
\unskip\
\newblock
\APACrefYearMonthDay{2017}{}{}.
\newblock
{\BBOQ}\APACrefatitle {Proximal policy optimization algorithms} {Proximal policy optimization algorithms}.{\BBCQ}
\newblock
\APACjournalVolNumPages{arXiv preprint arXiv:1707.06347}{}{}{}.
\PrintBackRefs{\CurrentBib}

\bibitem [\protect \citeauthoryear {%
Team%
\ \protect \BOthers {.}}{%
Team%
\ \protect \BOthers {.}}{%
{\protect \APACyear {2024}}%
}]{%
team2024gemini}
\APACinsertmetastar {%
team2024gemini}%
\begin{APACrefauthors}%
Team, G.%
, Georgiev, P.%
, Lei, V\BPBI I.%
, Burnell, R.%
, Bai, L.%
, Gulati, A.%
\BDBL {}others%
\end{APACrefauthors}%
\unskip\
\newblock
\APACrefYearMonthDay{2024}{}{}.
\newblock
{\BBOQ}\APACrefatitle {Gemini 1.5: Unlocking multimodal understanding across millions of tokens of context} {Gemini 1.5: Unlocking multimodal understanding across millions of tokens of context}.{\BBCQ}
\newblock
\APACjournalVolNumPages{arXiv preprint arXiv:2403.05530}{}{}{}.
\PrintBackRefs{\CurrentBib}

\bibitem [\protect \citeauthoryear {%
VanLehn%
}{%
VanLehn%
}{%
{\protect \APACyear {1990}}%
}]{%
vanlehn1990mind}
\APACinsertmetastar {%
vanlehn1990mind}%
\begin{APACrefauthors}%
VanLehn, K.%
\end{APACrefauthors}%
\unskip\
\newblock
\APACrefYear{1990}.
\newblock
\APACrefbtitle {Mind bugs: The origins of procedural misconceptions} {Mind bugs: The origins of procedural misconceptions}.
\newblock
\APACaddressPublisher{}{MIT press}.
\PrintBackRefs{\CurrentBib}

\bibitem [\protect \citeauthoryear {%
VanLehn%
, Ohlsson%
\BCBL {}\ \BBA {} Nason%
}{%
VanLehn%
\ \protect \BOthers {.}}{%
{\protect \APACyear {1994}}%
}]{%
vanlehn1994applications}
\APACinsertmetastar {%
vanlehn1994applications}%
\begin{APACrefauthors}%
VanLehn, K.%
, Ohlsson, S.%
\BCBL {}\ \BBA {} Nason, R.%
\end{APACrefauthors}%
\unskip\
\newblock
\APACrefYearMonthDay{1994}{}{}.
\newblock
{\BBOQ}\APACrefatitle {Applications of simulated students: An exploration} {Applications of simulated students: An exploration}.{\BBCQ}
\newblock
\APACjournalVolNumPages{Journal of artificial intelligence in education}{5}{}{135--135}.
\PrintBackRefs{\CurrentBib}

\bibitem [\protect \citeauthoryear {%
Weitekamp%
, Harpstead%
\BCBL {}\ \BBA {} Koedinger%
}{%
Weitekamp%
\ \protect \BOthers {.}}{%
{\protect \APACyear {2024}}%
}]{%
weitekamp2024ai2t}
\APACinsertmetastar {%
weitekamp2024ai2t}%
\begin{APACrefauthors}%
Weitekamp, D.%
, Harpstead, E.%
\BCBL {}\ \BBA {} Koedinger, K.%
\end{APACrefauthors}%
\unskip\
\newblock
\APACrefYearMonthDay{2024}{}{}.
\newblock
{\BBOQ}\APACrefatitle {AI2T: Building Trustable AI Tutors by Interactively Teaching a Self-Aware Learning Agent} {Ai2t: Building trustable ai tutors by interactively teaching a self-aware learning agent}.{\BBCQ}
\newblock
\APACjournalVolNumPages{arXiv preprint arXiv:2411.17924}{}{}{}.
\PrintBackRefs{\CurrentBib}

\bibitem [\protect \citeauthoryear {%
Weitekamp%
, Harpstead%
, MacLellan%
, Rachatasumrit%
\BCBL {}\ \BBA {} Koedinger%
}{%
Weitekamp%
\ \protect \BOthers {.}}{%
{\protect \APACyear {2019}}%
}]{%
weitekamp2019toward}
\APACinsertmetastar {%
weitekamp2019toward}%
\begin{APACrefauthors}%
Weitekamp, D.%
, Harpstead, E.%
, MacLellan, C\BPBI J.%
, Rachatasumrit, N.%
\BCBL {}\ \BBA {} Koedinger, K\BPBI R.%
\end{APACrefauthors}%
\unskip\
\newblock
\APACrefYearMonthDay{2019}{}{}.
\newblock
{\BBOQ}\APACrefatitle {Toward Near Zero-Parameter Prediction Using a Computational Model of Student Learning.} {Toward near zero-parameter prediction using a computational model of student learning.}{\BBCQ}
\newblock
\APACjournalVolNumPages{International Educational Data Mining Society}{}{}{}.
\PrintBackRefs{\CurrentBib}

\bibitem [\protect \citeauthoryear {%
Weitekamp%
, Rachatasumrit%
, Wei%
, Harpstead%
\BCBL {}\ \BBA {} Koedinger%
}{%
Weitekamp%
\ \protect \BOthers {.}}{%
{\protect \APACyear {2023}}%
}]{%
weitekamp2023simulating}
\APACinsertmetastar {%
weitekamp2023simulating}%
\begin{APACrefauthors}%
Weitekamp, D.%
, Rachatasumrit, N.%
, Wei, R.%
, Harpstead, E.%
\BCBL {}\ \BBA {} Koedinger, K.%
\end{APACrefauthors}%
\unskip\
\newblock
\APACrefYearMonthDay{2023}{}{}.
\newblock
{\BBOQ}\APACrefatitle {Simulating Learning from Language and Examples} {Simulating learning from language and examples}.{\BBCQ}
\newblock
\BIn{} \APACrefbtitle {International Conference on Artificial Intelligence in Education} {International conference on artificial intelligence in education}\ (\BPGS\ 580--586).
\PrintBackRefs{\CurrentBib}

\bibitem [\protect \citeauthoryear {%
Weitekamp%
, Ye%
, Rachatasumrit%
, Harpstead%
\BCBL {}\ \BBA {} Koedinger%
}{%
Weitekamp%
\ \protect \BOthers {.}}{%
{\protect \APACyear {2020}}%
}]{%
weitekamp2020investigating}
\APACinsertmetastar {%
weitekamp2020investigating}%
\begin{APACrefauthors}%
Weitekamp, D.%
, Ye, Z.%
, Rachatasumrit, N.%
, Harpstead, E.%
\BCBL {}\ \BBA {} Koedinger, K.%
\end{APACrefauthors}%
\unskip\
\newblock
\APACrefYearMonthDay{2020}{}{}.
\newblock
{\BBOQ}\APACrefatitle {Investigating Differential Error Types Between Human and Simulated Learners} {Investigating differential error types between human and simulated learners}.{\BBCQ}
\newblock
\BIn{} \APACrefbtitle {International Conference on Artificial Intelligence in Education} {International conference on artificial intelligence in education}\ (\BPGS\ 586--597).
\PrintBackRefs{\CurrentBib}

\bibitem [\protect \citeauthoryear {%
Xiao%
\ \BBA {} Zhang%
}{%
Xiao%
\ \BBA {} Zhang%
}{%
{\protect \APACyear {2023}}%
}]{%
xiao2023deep}
\APACinsertmetastar {%
xiao2023deep}%
\begin{APACrefauthors}%
Xiao, Z.%
\BCBT {}\ \BBA {} Zhang, D.%
\end{APACrefauthors}%
\unskip\
\newblock
\APACrefYearMonthDay{2023}{}{}.
\newblock
{\BBOQ}\APACrefatitle {A deep reinforcement learning agent for geometry online tutoring} {A deep reinforcement learning agent for geometry online tutoring}.{\BBCQ}
\newblock
\APACjournalVolNumPages{Knowledge and Information Systems}{65}{4}{1611--1625}.
\PrintBackRefs{\CurrentBib}

\end{thebibliography}

\end{document}